\documentclass{egpubl}
\usepackage{pg2023s}

\WsShortPaper      

\usepackage[T1]{fontenc}
\usepackage{dfadobe}  

\usepackage{cite}  
\BibtexOrBiblatex
\electronicVersion
\PrintedOrElectronic
\ifpdf \usepackage[pdftex]{graphicx} \pdfcompresslevel=9
\else \usepackage[dvips]{graphicx} \fi

\usepackage{egweblnk} 

\usepackage{amsmath}
\usepackage{amssymb}
\usepackage{times}  
\usepackage{multirow}
\usepackage{color}

\graphicspath{{./pdf/}{./img/}}

\title[SketchBodyNet: A Sketch-Driven Multi-faceted Decoder Network for 3D Human Reconstruction]%
      {SketchBodyNet: A Sketch-Driven Multi-faceted Decoder Network for 3D Human Reconstruction}

\author[Wang et al.]
{\parbox{\textwidth}{\centering Fei Wang$^{1}$\orcid{0000-0001-8949-1894},~~Kongzhang Tang$^{1}$\orcid{0000-0002-1105-8449},~~Hefeng Wu$^{2}\,$\thanks{Corresponding author: Hefeng Wu (wuhefeng@mail.sysu.edu.cn)}\orcid{0000-0002-2132-6515},~~Baoquan Zhao$^{2}$\orcid{0000-0002-0574-1663},~~Hao Cai$^1$,~~and~~Teng Zhou$^3$\orcid{0000-0003-1920-8891} 
        }
        \\
{\parbox{\textwidth}{\centering $^1$ Shantou University \quad\quad
         $^2$ Sun Yat-sen University \quad\quad
        $^3$ Hainan University \vspace{0.1ex}\\ \,
       }
}
}

\begin{document}


\maketitle
\begin{abstract}
    Reconstructing 3D human shapes from 2D images has received increasing attention recently due to its fundamental support for many high-level 3D applications. Compared with natural images, freehand sketches are much more flexible to depict various shapes, providing a high potential and valuable way for 3D human reconstruction. However, such a task is highly challenging. 
    The sparse abstract characteristics of sketches add severe difficulties, such as arbitrariness, inaccuracy, and lacking image details, to the already badly ill-posed problem of 2D-to-3D reconstruction.
    Although current methods have achieved great success in reconstructing 3D human bodies from a single-view image, they do not work well on freehand sketches.
    In this paper, we propose a novel sketch-driven multi-faceted decoder network termed SketchBodyNet to address this task. 
    Specifically, the network consists of a backbone and three separate attention decoder branches, where a multi-head self-attention module is exploited in each decoder to obtain enhanced features, followed by a multi-layer perceptron. The multi-faceted decoders aim to predict the camera, shape, and pose parameters, respectively, which are then associated with the SMPL model to reconstruct the corresponding 3D human mesh. In learning, existing 3D meshes are projected via the camera parameters into 2D synthetic sketches with joints, which are combined with the freehand sketches to optimize the model.
    To verify our method, we collect a large-scale dataset of about 26k freehand sketches and their corresponding 3D meshes containing various poses of human bodies from 14 different angles.
    Extensive experimental results demonstrate our SketchBodyNet achieves superior performance in reconstructing 3D human meshes from freehand sketches.


\end{abstract}

\section{Introduction}

\begin{figure}[t!]
	\centering
	\begin{minipage}{0.32\linewidth}
		\centering
		\includegraphics[width=1\textwidth]{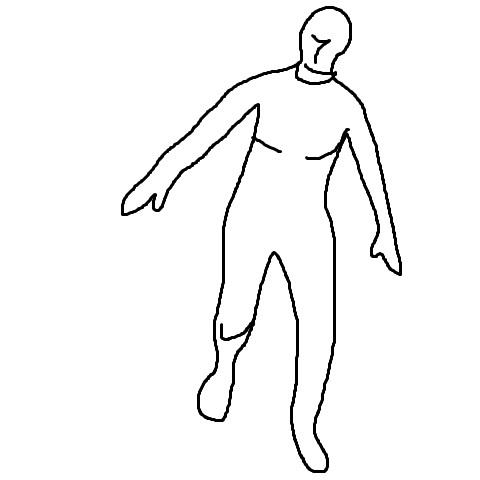} \\
		\includegraphics[width=1\textwidth]{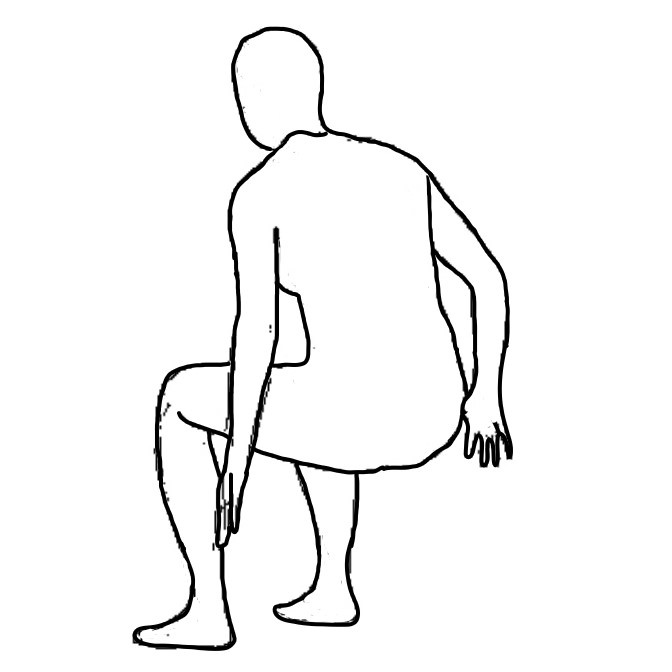}
	\end{minipage}
	\hfill
	\begin{minipage}{0.32\linewidth}
		\centering
		\includegraphics[width=1\textwidth]{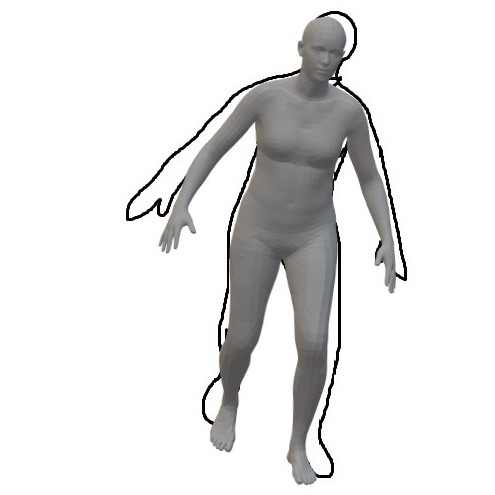} \\
		\includegraphics[width=1\textwidth]{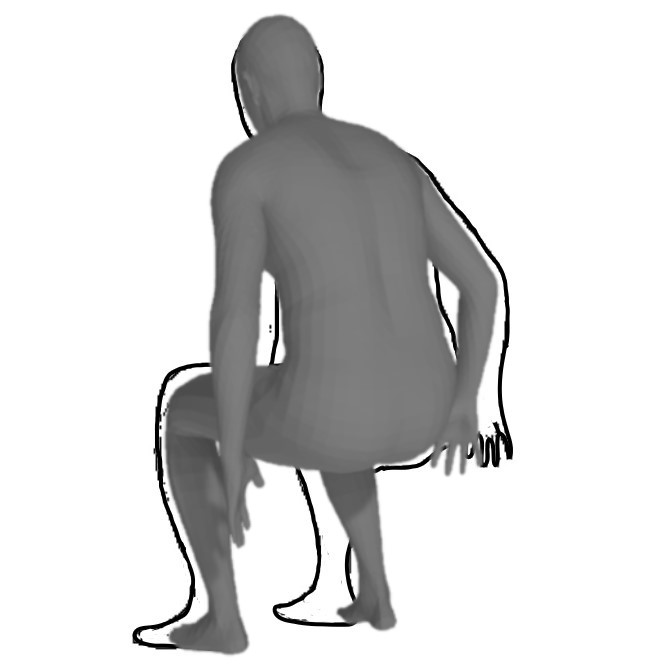}
	\end{minipage}
	\hfill
	\begin{minipage}{0.32\linewidth}
		\centering
		\includegraphics[width=1\textwidth]{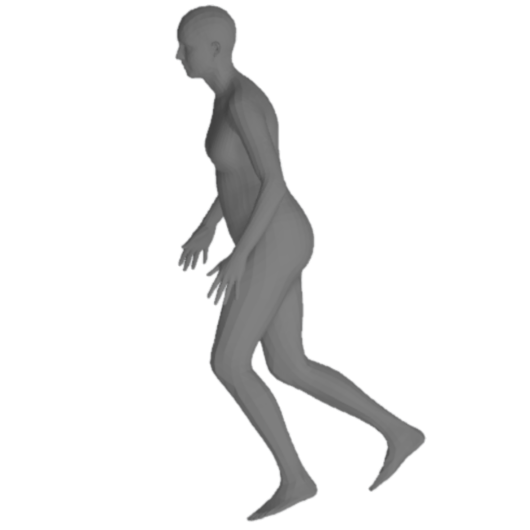} \\
		\includegraphics[width=1\textwidth]{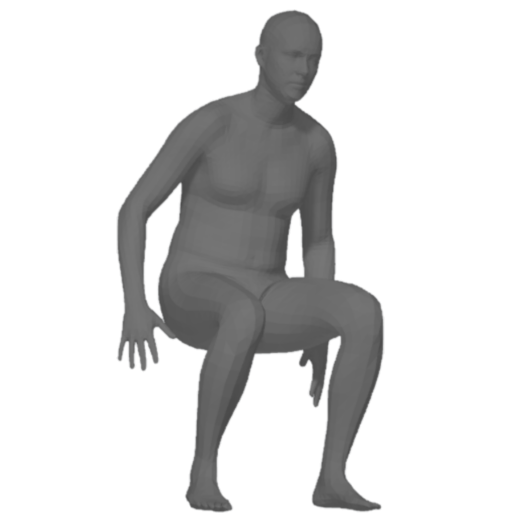}
	\end{minipage}
	\caption{Illustration of our sketch-driven 3D human modeling. The 3D human mesh (Right) is directly reconstructed from sketches (Left). It can be observed that the 2D projection of the 3D mesh aligns well with the sketch (Middle). Our method is suitable for users of different drawing levels. The more standard the sketch tends to be, the more accurate the reconstruction and projection will be.}
	\label{fig1}
\end{figure}

A sketch is a simple and convenient way of expressing ideas \cite{WangLWLWLH19icme}. 
We use sketches flexibly for the transient expression of the objects depicted \cite{LiWHLWL17icme}. 
With the popularity of mobile phones, tablets, computers, and other electronic devices, it is easy and efficient for users to draw a sketch. 
It provides a high potential and valuable way for 2D-to-3D human reconstruction, as illustrated in Fig. \ref{fig1}, since directly generating 3D human meshes from sketching can dramatically improve the working efficiency of designers in modeling tasks.
In conventional procedures, designers often first draw sketches for their idea before the fine modeling in conditions of meeting the satisfaction of customers.
It often takes less than ten seconds to sketch a human body, but a lot of time to manually construct the corresponding 3D human body. 
Moreover, the modeling process generally requires professional software and expert skills with plenty of study time. 
These increasing demands of 3D human modeling motivate the development of fast sketch-based 3D modeling pipelines.
Therefore, this work aims to model a 3D mannequin directly from a few simple strokes without requiring users' professional drawing and modeling skills.
professional drawing and modeling skills. 

However, it is highly challenging to model mannequins directly from sketches, because such a single-view 3D reconstruction task is badly ill-posed and the sparse abstract characteristics of sketches pose more difficulties, such as the arbitrariness and inaccuracy of strokes, the lack of texture and color information, etc.
Although reconstructing 3D shapes from 2D monocular images has made great progress, the off-the-self 3D reconstruction methods do not work well on this task for the following reasons.
1) This kind of method often aims at natural images with rich texture, which requires a large number of photos of humans or mannequins with different poses, often necessary to be taken professionally to meet specific conditions. 
2) These methods may not learn to capture the key information of the sketch, challenged by various difficulties like the confusion of lines, the imbalance of shape proportion, and the unforeseeable sketch drawing by different users.
3) Previous methods generally require 2D annotations of the natural images, e.g., 2D marking joints of humans, to obtain an accurate projection of the 3D human body mesh in the 2D space, but there are no such annotations provided in the freehand sketches.

\begin{figure*}[t!]
	\centering
	\includegraphics[width=1\textwidth]{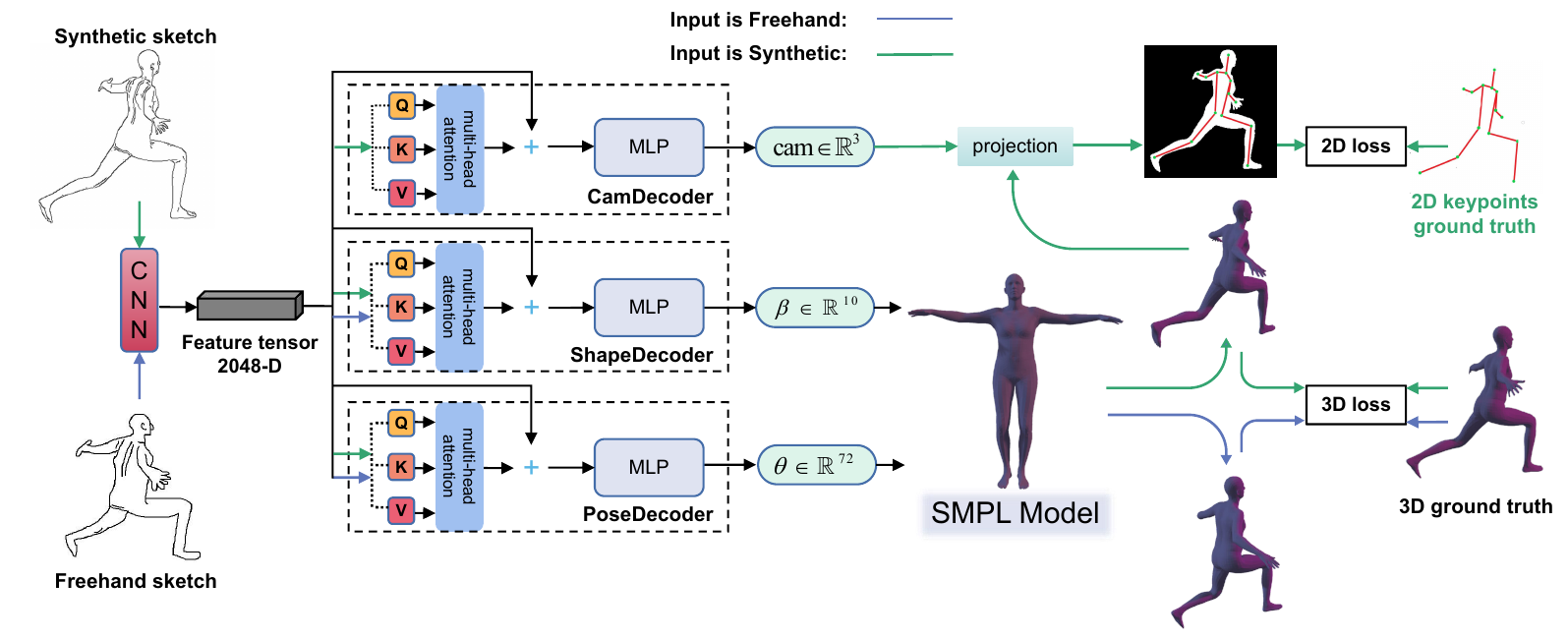}
	\caption{Our network consists of a backbone CNN and three separate attention decoder branches, with each branch in charge of a specific facet, denoted as CamDecoder, ShapeDecoder and PoseDecoder, respectively. We also introduce a  multi-head self-attention module in the decoder structure for feature enhancement. The predicted shape and pose parameters $\beta$ and $\theta$ are input into the SMPL model to generate the 3D mesh of the human body, and the camera parameter $cam$ can help project the 3D mesh onto a 2D plane.
		\label{fig2}}
\end{figure*}

To address these issues, we propose a  sketch-driven multi-faceted decoder network architecture and a new learning pipeline to reconstruct 3D human mesh from the sketch, as illustrated in  Fig.~\ref{fig2}. 
The key idea is to capture the characteristics of sketches through an improved encoder-decoder structure.
In the encoding stage of our network, we follow previous work to use deep residual networks (e.g., ResNet50) for feature extraction. 
While in the decoding stage, all existing works use a single decoder to predict the camera, shape and pose parameters jointly. 
We argue that such configuration may induce these parameters to generate a negative influence on each other. 
In contrast, we devise a new multi-faceted network structure that uses three separate attention decoder branches to predict the camera, shape and pose parameters, respectively, so as to reduce the negative mutual influence between different parameters.
During learning, we lock the parameters of individual branches at each time to avoid conflicts between different branches. 
Moreover, we use a multi-head self-attention module in each decoder branch, which leads the branch to focus on the key positions of the sketch, such as the key line of the sketch.
In this way, the model can better learn to utilize the key information in the sketch. 
In order to obtain the mapping relationship between the 3D and 2D key points of the human body, we project existing 3D meshes into 2D synthetic sketches with 2D joints and learn the parameter space of potential projection for the freehand sketches. 
By combining synthetic and freehand sketches, our new learning pipeline can take full advantage of the two-dimensional information about the human body.
In the inference phase, our method automatically generates a 3D human mesh from a simple freehand sketch and can roughly project the 3D mesh onto the position of the human body in the sketch. 
Our method achieves good reconstruction performance for both simple and fine sketch depictions of the human body pose.

Since this task is hardly studied in the literature,
we further build a large-scale freehand sketches dataset of various styles, termed Sketch3D, to evaluate the performance of our method.
By referring to existing human pose datasets ~\cite{ionescu2013human3,mehta2017monocular,johnson2010clustered,lassner2017unite}, we collect a total of 26k freehand sketches of various styles for the new Sketch3D dataset. 
Our dataset is characterized by the most diversity of postures compared with previous datasets, and it is sufficient in quantity for deep learning models. 
We also generate synthetic sketches from existing 3D human meshes.
These synthetic sketches can be regarded as sketches of human bodies drawn by professionals. 
Extensive experiments are conducted on this dataset to evaluate our method, demonstrating that it can effectively generate a 3D human body mesh from freehand sketches and achieve superior performance over existing leading methods. 

The main contributions of our paper can be summarized as follows.
\begin{itemize}
	\item We investigate a sketch-to-3D human reconstruction task and propose a multi-faceted attention decoder network SketchBodyNet. Its architecture is distinct from existing works and more effective, and we devise a new learning pipeline for this task.
	\item We build a large-scale dataset Sketch3D, which contains 26k freehand sketches of various styles and features the most diversity of human postures.
	\item We conduct extensive experiments and the results show that our method outperforms state-of-the-art 3D human reconstruction models.
\end{itemize}

\section{Related work}
	
\subsection{Single-View 3D Human Reconstruction}

One pipeline for reconstructing the 3D human body from a single-view natural image is to first predict the 3D pose directly from the 2D pose of the human body and then use the 3D pose to obtain the 3D shape. 
Benefiting from the rapid advancements of deep neural networks \cite{he2016deep,vaswani2017attention,WuHWLNC19pr,ChenLCHW22pami,chen2022cross}, there are a lot of previous works devoted to regressing 3D joints and shapes from 2D joints.
Martinez \emph{et al.} \cite{martinez2017simple} use a simple neural network to directly transform the pose from 2D space to 3D space and obtain impressive performance. 
Choi \emph{et al.} \cite{choi2020pose2mesh} present a network fusion method by fusing PostNet and MeshNet together to reconstruct 3D mesh. PoseFormer ~\cite{zheng20213d} is a pose regressor based on Transformer ~\cite{vaswani2017attention}, which uses Transformer to transform 2D joint nodes to 3D poses. 

These previous methods have been gradually replaced by recent end-to-end methods based on the SMPL model \cite{loper2015smpl}. Current leading methods generally directly regress SMPL parameters from a single RGB image for 3D reconstruction.
Lassner \emph{et al.} \cite{lassner2017unite} acquire a lot of high-quality 3D poses from 2D poses by introducing SMPLify \cite{2016Keep}.
Varol \emph{et al.} \cite{varol2018bodynet} obtain 3D mannequins directly from single images.
There are also plenty of methods for improving the data and networks on top of end-to-end training.
Kolotouros \emph{et al.} \cite{kolotouros2019learning} use SMPLify \cite{2016Keep} to monitor the training of the regression network so that the model can converge faster. 
Sengupta \emph{et al.} \cite{sengupta2020synthetic} reconstruct the 3D shape from the image by synthetic training.
Moon \emph{et al.} \cite{moon2020i2l} reconstruct the 3D grid using the heat map pixels of the image as input.
Zhang \emph{et al.} \cite{zhang2021pymaf} incorporate the subsampled features of the 3D pose into the pose regression network for multi-period prediction.
Dwivedi \emph{et al.} \cite{dwivedi2021learning} propose a new differentiable semantic loss, which divides the human body into non-occluded and occluded parts to calculate different forms of losses separately.
Li \emph{et al.} \cite{li2021hybrik} propose a joint rotation method based on 3D pose regression to learn from images how to distort and rotate 3D models.
Similar work can be found in \cite{wan2021encoder}.
There also are some methods that directly predict the three-dimensional mesh from RGB images~\cite{kolotouros2019convolutional}.

\subsection{Sketch-Based 3D Reconstruction}
Sketches are a highly abstract but very convenient way of expression, and they are used commonly in our work and daily life. 
The sketch is employed in many computer-aided tasks, such as corruption repair ~\cite{qi2021sketchlattice,GaoLWW17spic}, synthesis ~\cite{li2017free}, image retrieval ~\cite{eitz2010sketch,DengWWHLL18ivc}, segmentation ~\cite{wang2020multi}, scene reconstruction ~\cite{shin2007magic}, 3D shape retrieval \cite{WangLWWL16siggraph,WangLLWWZ17cgf}, etc. 
For 3D reconstruction from a freehand sketch, There are some works trying to solve this problem. 
Wang \emph{et al.} \cite{wang20203d} and Luo \emph{et al.} \cite{luo2021simpmodeling} rebuild the 3D model of animal heads by using neural networks.
Zhang \emph{et al.} \cite{zhang2021sketch2model} use sketches to predict 3D meshes of some rigid objects, such as cars, tables, chairs, airplanes, etc.
However, these works do not involve the sketch of the human body, and the parameter space of the human body is much different from those of these objects.
There are only several works that try to estimate 3D human pose via sketch.
Brodt and Bessmeltsev ~\cite{brodt2022sketch2pose} transfer the sketch pose to the standardized 3D mannequin model by first predicting three key elements (i.e., 2D bone tangents, self-contacts, and foreshortening) and then using them in a nonlinear optimization process to produce the final result, which is complex and requires online optimization for each input.
Unlu \emph{et al.} ~\cite{unluinteractive} exploit interactive sketching to build a 3D mannequin pose by drawing each part of the human sketch as a cylinder.
In this work, we aim to learn an efficient model that can reconstruct the accurate 3D human body mesh directly from the freehand sketch without any optimization in the inference stage. 

\section{Method}
In this section, we first describe the parametric 3D human body model SMPL~\cite{loper2015smpl} and then present the proposed method in detail.

\subsection{SMPL Model}
Our method builds on a parametric 3D human model, i.e.,  the Skinned Multi-Person Linear (SMPL) model ~\cite{loper2015smpl}. 
The SMPL model can represent different poses and shapes of the human body by changing the corresponding parameters.
It can simulate the bulge and depression of human muscles and other tissues in the process of limb movement, so as to carry out 3D human modeling and animation in any pose. 
The SMPL model takes the shape parameter $\beta$ and the pose parameter $\theta$ as input, and outputs a 3D human body model with 6,890 mesh vertices and 13,776 triangular mesh surfaces, and a 3D skeleton with parent-child structures.
The whole human body model has a total of 23 different joints, each joint has three degrees of freedom, plus three variables to control the spatial position of the SMPL model, so the pose parameter $\theta$ has 23 $\times$ 3 + 3 = 72 dimensions. The body shape parameter $\beta$ is a feature vector with a length of 10, each dimension represents a shape basis, and it can control the degree of height, fatness, and muscle lines of the human body. For the parametric SMPL model, we follow the formulation of Loper \emph{et al.}~\cite{loper2015smpl}, which is defined as:
\begin{equation}
    \begin{aligned}
    	T(\beta,\theta)=T_{0}+B_{s}(\beta)+B_{p}(\theta),
     \end{aligned}
\end{equation}
\begin{equation}
    \begin{aligned}
    	M(\beta,\theta)=W(T(\beta,\theta),J(\beta), \theta,\beta),
     \end{aligned}
\end{equation}
where $T_0$ denotes the mean template shape with zero pose, $B_s(\beta)$ and $B_p(\theta)$ are the body deformation functions, $J(\beta)$ predicts the joint locations, and $W(\cdot)$ is a blend skinning function. 

\subsection{SketchBodyNet Architecture}
The architecture of our SketchBodyNet is illustrated in Fig. \ref{fig2}. 
It mainly includes a backbone convolutional neural network (CNN) and three separate attention decoder branches, which will be detailed in the following.

\noindent\textbf{Backbone Network.}
We follow previous works to use a deep residual network ~\cite{he2016deep} as the backbone, which is well-validated in the literature. It may preserve the original image features to a greater extent for sketches. We use ResNet50 to extract image features whose input size is $3\times224\times224$, and it maps the input image to a 2048-D feature vector.
Specifically, Given an input image $I\in\mathbb{R}^{224\times224\times3}$, the backbone network first obtains an image feature $f_{image}\in\mathbb{R}^{7\times7\times2048}$ and then pools it into 2048-D feature vector $f_{in}$.

\noindent\textbf{Multi-faceted Decoders.}
Previous methods all use one branch to predict all parameters of pose, shape or camera projection together, which may not optimize well due to strong mutual interference. 
In order to make these three parameters do not affect each other negatively, we propose a multi-faceted attention decoder architecture based on the self-attention mechanism \cite{vaswani2017attention}, which makes the attention mechanism work on multiple decoding branches independently. Our multi-faceted decoding structure consists of three decoder branches (i.e., PoseDecoder, ShapeDecoder, CamDecoder). 
All decoder branches share the same architecture, which comprises a multi-head self-attention module and a multi-layer perception (MLP) module, as shown in Fig. \ref{fig2}.
Each branch takes the feature vector extracted by the backbone network as input and use independent self-attention to predict the pose, shape, and camera parameters, respectively.
Extensive evaluation show that our design can outperform previous methods favorably.

\noindent\textbf{Multi-head Self-attention Module.}
The design of this module is as follows. Given the 2048-D feature vector $f_{in}$ extracted by backbone network, we follows a similar self-attention mechanism used in Transformer  \cite{vaswani2017attention,YuanCWWC22cee}.
To be specific, we use the feature vector $f_{in}$ to obtain a query vector $Q\in\mathbb{R}^{8\times1\times64}$, a key vector $K\in\mathbb{R}^{8\times1\times64}$ and a value vector $V\in\mathbb{R}^{8\times1\times64}$ through a linear layer. Then they are computed with the multi-head self-attention mechanism to produce a globally enhanced feature for each branch, i.e.,  $f_*^{a}\in\mathbb{R}^{2048}$, $*\in\{pose, shape, cam\}$. Finally, $f_{in}$ and $f_*^{a}$ are added to obtain the optimized global feature $x_* \in\mathbb{R}^{2048}$ for an independent branch. The procedure is formulated as:
\begin{equation}
	\begin{aligned}
		&Q_{*}, K_{*}, V_{*}=Linear_{*}(f_{in}),
	\end{aligned}
\end{equation}
\begin{equation}
	\begin{aligned}
		&f_{*}^{a}=Attention_{*}(Q_{*},K_{*},V_{*}),
	\end{aligned}
\end{equation}
\begin{equation}
	\begin{aligned}
		&x_{*}=f_{in} + f_{*}^{a}, \quad *\in\{pose, shape, cam\}.
	\end{aligned}
\end{equation}

\noindent\textbf{Prediction.}
After the above procedure, we obtain the optimized global features $x_{pose}\in\mathbb{R}^{2048}$, $x_{shape}\in\mathbb{R}^{2048}$, $x_{cam}\in\mathbb{R}^{2048}$ for the pose, shape and camera projection branches, respectively. Then we perform concatenation operation with the initial parameters $\theta_{init}\in\mathbb{R}^{72}$, $\beta_{init}\in\mathbb{R}^{10}$ and $cam_{init}\in\mathbb{R}^{3}$, which are corresponding to the SMPL initialization model, and input them to separate linear projection layers in each branch to get the final results,  i.e.,
\begin{equation}
	\begin{aligned}
		&\theta=PoseDecoder(Concat(x_{pose}, \theta_{init})),
	\end{aligned}
\end{equation}
\begin{equation}
	\begin{aligned}
		&\beta=ShapeDecoder(Concat(x_{shape}, \beta_{init})),
	\end{aligned}
\end{equation}
\begin{equation}
	\begin{aligned}
		& cam=CamDecoder(Concat(x_{cam}, cam_{init})).
	\end{aligned}
\end{equation}
where $\theta_{init}$, $\beta_{init}$, and $cam_{init}$ are SMPL initialization parameters, and $\theta\in\mathbb{R}^{72}$, $\beta\in\mathbb{R}^{10}$, $cam\in\mathbb{R}^{3}$ are the final predicted parameters for pose, shape and camera projection, respectively.

Afterwards, the predicted pose and shape parameters (i.e., $\theta$ and $\beta$)  are passed to SMPL to produce 3D grid vertices of human body. The 3D grid vertices can then be used with  the predicted camera projection parameter $cam$ to produce a 2D rendering.

\subsection{Learning Pipeline}
Our training process is a step-wise refinement training (SRT) process.
First, the model is trained with the synthetic sketch data for some rounds. In this way, the model can learn a good initialization at the synthetic data since they contain more information, i.e., the camera parameters and the 2D ground truth keypoints projected from the 3D human body. Then, the model is trained for some more rounds on the freehand sketch data and finally obtain our target model.

Due to the difference in 2D ground truth between the freehand sketch and the synthetic sketch, we distinguish the synthetic sketch and the freehand sketch in the back-propagation process. If the synthetic sketch is used as input, we take 3D shapes and 2D keypoints as supervision; If the freehand sketch is used as input, we only use the 3D pose as supervision. The loss of each part is defined as follows: Firstly, the ground truth grid vertices of the 3D body are defined as $M_{3D}$, and the predicted grid vertices are defined as $\hat{M}_{3D}=M(\hat{\beta},\hat{\theta})$. The ground truth 3D joints are $V_{3D}$ and predicted 3D joints are $\hat{V}_{3D}$. The ground truth 2D joints are $V_{2D}$, and the predicted 2D joints are $\hat{V}_{2D}$, which are obtained by projecting 3D joints onto the 2D plane using the predicted camera parameters $cam\in\mathbb{R}^{3}$.

We use $\mathcal{L}1$ Loss to calculate the error between the predicted and ground-truth grid vertices:
\begin{equation}
    \begin{aligned}
	   \mathcal{L}_{shape_{3D}}=\sum_{i=1}^{N}\Vert\hat{M}^i_{3D}-M^i_{3D}\Vert_1.
    \end{aligned}
\end{equation}
We use $\textrm{MSE}$ Loss for measuring the prediction of 3D and 2D joints:
\begin{equation}
    	\mathcal{L}_{3D}=\frac{\sum_{i=1}^{N}(\hat{V}^i_{3D}-V^i_{3D})^{2}}{N}, \quad 
         	\mathcal{L}_{2D}=\frac{\sum_{i=1}^{N}(\hat{V}^i_{2D}-V^i_{2D})^{2}}{N}.
\end{equation}
As for the prediction of the SMPL parameters, the loss is defined as:
\begin{equation}	
    	\mathcal{L}_{\theta}=\frac{\sum_{i=1}^{N}(\hat{\theta}^i-\theta^i)^{2}}{N}, \quad
    	\mathcal{L}_{\beta}=\frac{\sum_{i=1}^{N}(\hat{\beta}^i-\beta^i)^{2}}{N}.
\end{equation}

To sum up, if synthetic sketches are used in training our network, we can use 2D joints and shape information as supervisors, and our losses are as follows:
\begin{equation}	
	\begin{aligned}
        \mathcal{L}_1=\mathcal{L}_{shape_{3D}}+\mathcal{L}_{3D}+\mathcal{L}_{2D}+\mathcal{L}_{\theta}+\mathcal{L}_{\beta}.
    \end{aligned}
\end{equation}

If freehand sketches are used in training our network, we only used the pose parameter $\theta$ of SMPL and the 3D joints of the final 3D body as the supervision for training. The losses are as follows:
\begin{equation}	
    \begin{aligned}
    	\mathcal{L}_2=\mathcal{L}_{3D}+\mathcal{L}_{\theta}.
     \end{aligned}
\end{equation}

\section{Sketch3D Dataset Construction}
In this section, we mainly introduce the collection and specific content of our dataset. The 3D source data of our dataset is mainly from the three open-source data sets of MPI-INF-3DHP, UP-3D and LSP.

The {MPI-INF-3DHP}~\cite{mehta2017monocular} dataset contains information about various human postures. 
We use both synthetic and hand-drawn sketch data from this dataset for training and evaluation.

The {LSP}~\cite{johnson2010clustered} dataset in its original task mainly serves as a validation of the performance of image-model alignment.
We use both synthetic and hand-drawn sketch data from this dataset to validate the 3D model projection alignment.

The {UP-3D}~\cite{lassner2017unite} is only available as a supplement to the training set.
We use the synthetic sketches of this dataset as a training supplement for the alignment of the 3D model projection.

Our dataset contains various poses of the human body from 14 different angles.
We recruited 41 students as drawing volunteers and asked each student to draw about 600 pictures.
First, we provided a small training session for the volunteers.
we put forward different requirements for them: (1) Volunteers with painting skills had better be able to restore the details of the human body from the texture map as much as possible, including the details of the face and the key folds of projection; (2) For the volunteers with average painting skills, we asked them to roughly restore the information on the edge of the human body;
(3) For the volunteers with poor drawing skills, we only asked them to draw the general posture outline of the human body. 
Then, we asked the volunteers to use the drawing board of Windows system to draw sketches that only contain black and white colors. 

\begin{figure}[t!]
	\centering
	\includegraphics[width=0.5\textwidth]{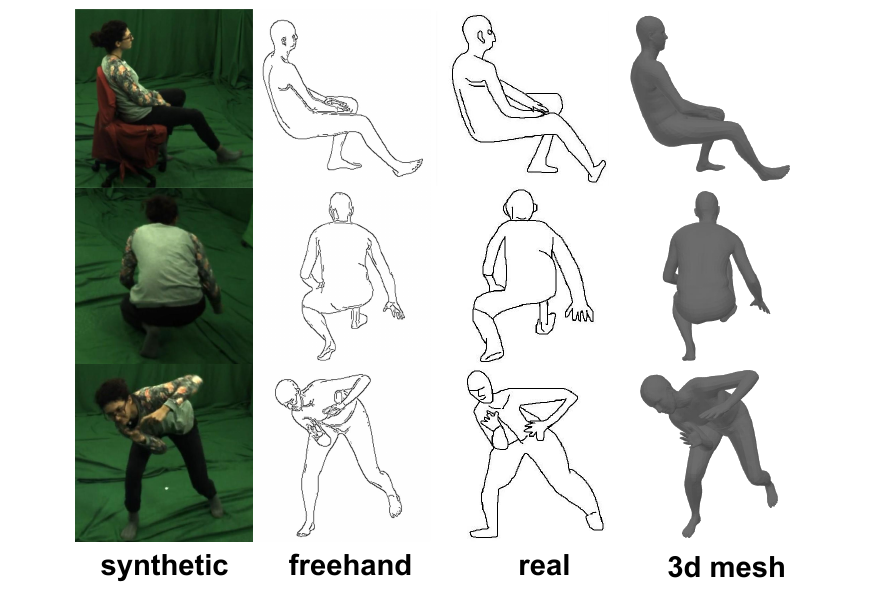}
	\caption{\centering{Example illustration of our dataset. 
 }
		\label{fig3}}
\end{figure}

Through data collection and collation, we get a new dataset consisting of freehand sketches and synthetic sketches, as shown in Fig.~\ref{fig3}. 
Synthetic sketches are images obtained by projecting the original 3D model at a specific angle onto a 2D plane and extracting the edge lines using OpenCV. 

We named the new dataset Sketch3D, which has about 26k map sketches and their corresponding 3D models, making it the largest sketch dataset available. Some examples are exhibited in Fig.~\ref{fig3}.

\section{Experimental evaluation}

This section focuses on the implementation details and extensive evaluation of the proposed method.

\subsection{Evaluation Metrics}

For a comprehensive comparison, we use four evaluation metrics, \emph{i.e.}, MPJPE, Reconst.Error, \emph{Acc.}, and \emph{F1}, to evaluate the performance of our method, which are widely used in previous studies~\cite{kolotouros2019convolutional,kanazawa2018end,kolotouros2019learning,wan2021encoder}.

\textbf{MPJPE} is the Mean Per Joint Position Error, which is defined  in~\cite{ionescu2013human3}:
\begin{equation}
	E_{MPJPE}(f,\mathcal{S})=\frac{1}{N_{\mathcal{S}}}\sum_{i=1}^{N_{\mathcal{S}}}\Vert{m^{(f)}_{\textbf{f},\mathcal{S}}(i)-m^{(f)}_{\textbf{gt},\mathcal{S}}(i)}\Vert_2.
\end{equation}

\textbf{Reconst.Error} is the Reconstruction Error, which is the MPJPE absolutely aligning the predicted result with the ground truth after Procrustes Analysis~\cite{kolotouros2019learning}. 

\textbf{\emph{Acc.}} is the image-model alignment accuracy.
The \emph{Acc.} indicates the alignment quasi-energy rate after projecting a three-dimensional grid onto a two-dimensional plane without three-dimensional data~\cite{johnson2010clustered}. 
For our sketch, we only need to fill the interior of the sketch and perform intersection and comparison calculations on the two-dimensional projection. 

\textbf{\emph{F1}} is the \emph{F1} score~\cite{kolotouros2019learning}, which is an index measuring the accuracy of binary models in statistics. 
It takes the precision and recall rate of the classification model into account.

\begin{figure*}[t!]
	\centering
	\includegraphics[width=0.85\linewidth]{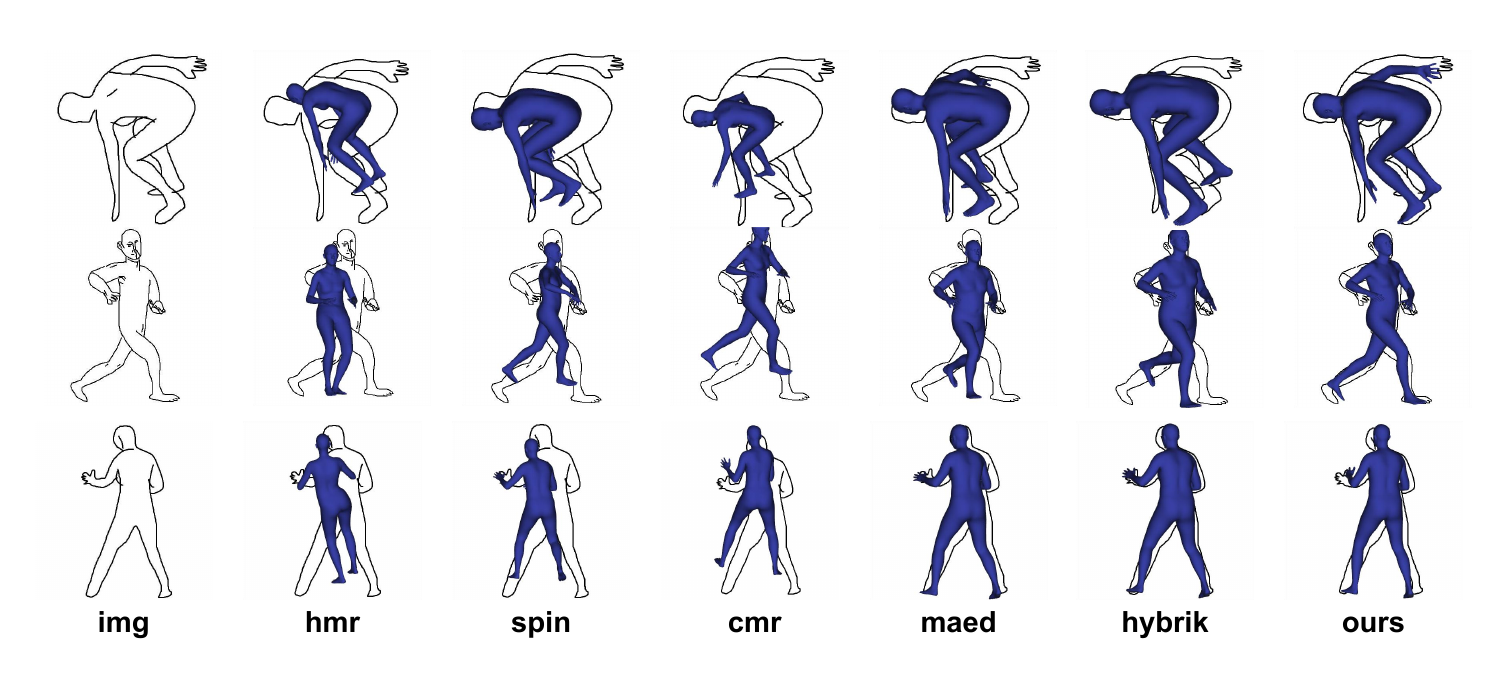}
	\caption{\centering{Qualitative comparisons on the Sketch3D dataset. 
 }
		\label{fig4}}
\end{figure*}

\begin{table*}[t!]
	\centering
	\caption{Comparison with the state-of-art methods on the image-model alignment and 3D Reconstruction Error on freehand data and synthetic data. 
 }
	\label{tab1}
	\begin{tabular}{c|cccc|cccc}
		\hline
		\multirow{2}{*}{Model}&\multicolumn{4}{c|}{Synthetic sketch}&\multicolumn{4}{c}{Freehand sketch} \\ \cline{2-9}
		&MPJPE$\downarrow$&Reconst. Error$\downarrow$&$\mathit{Acc.}$$\uparrow$&$\mathit{F1}$$\uparrow$
		&MPJPE$\downarrow$&Reconst. Error$\downarrow$&$\mathit{Acc.}$$\uparrow$&$\mathit{F1}$$\uparrow$ \\
		\hline 
		HMR[2018]&201.11&115.70&85.15&76.67&240.95&118.36&82.66&73.54 \\
		SPIN[2019]&133.55&84.54&84.72&73.43&185.09&99.98&82.29&70.71 \\
		CMR[2019]&136.98&85.15&82.36&69.67&202.26&101.48&80.03&66.81 \\
		MAED[2021]&125.73&81.10&87.32&78.00&176.79&97.24&85.07&75.35 \\
            Hybrik[2023]&160.71&105.83&77.92&43.79&188.95&109.97&76.00&43.18 \\
		\hline
		Ours&\textbf{123.95}&\textbf{80.39}&\textbf{89.28}&\textbf{82.10}&\textbf{154.89}&\textbf{91.19}&\textbf{86.93}&\textbf{79.23}\\
		\hline
	\end{tabular}
\end{table*}

\subsection{Implementation Details}

Due to the different drawing habits of different users, the sketch is not always in the middle of the image. 
Previous methods detect the position of the human body by marking the human body from 2D joints. 
However, the freehand sketch does not have 2D joints. 
To solve this problem, we use a preprocessor to directly detect the bounding box of the human body in the sketch. After that, the bounding box is used to segment and crop the human body in the sketch image. 
In this way, we get a well-cropped, accurately positioned sketch of the human body. 
We directly use the cropped image ($\mathbb{R}^{224\times224\times3}$) of the sketch as the input to our network.
Although the human body sketches are in black lines and white background, we store them in the RGB format to facilitate compatibility with the color image for convenient processing.

We train our network on NVIDIA RTX2080TI GPU. The learning rate is set to $1\mathit{e}-4$, and the batch size is $64$.
In our approach, synthetic data and freehand data are used in training step-wisely. For SRT, we firstly use synthetic sketch images with 2D and 3D data to train our network for 100 epochs. Then, we use freehand sketch images with only 3D data to train for 10 epochs.

\begin{table*}[t!]
	\centering
	\caption{Ablation Study on Dataset and Attention.}
	\label{tab2}
	\begin{tabular}{c|cc|cccc|cccc}
		\hline
		Model&\multicolumn{2}{c|}{Train dataset}&\multicolumn{4}{c|}{Synthetic sketch}&\multicolumn{4}{c}{Freehand sketch} \\
		\hline
		Attention&Syn.&Free.&MPJPE$\downarrow$&Reconst. Error$\downarrow$&$\mathit{Acc.}$$\uparrow$&$\mathit{F1}$$\uparrow$&MPJPE$\downarrow$&Reconst. Error$\downarrow$&$\mathit{Acc.}$$\uparrow$&$\mathit{F1}$$\uparrow$ \\
		\hline
		\multirow{2}{*}{}&\checkmark&&139.80&93.32&87.60&80.67&231.31&103.38&85.55&78.33 \\
		&&\checkmark&135.48&86.72&77.92&43.18&160.09&94.01&75.99&43.18\\	&\checkmark&\checkmark&149.33&92.69&85.29&73.92&208.20&102.37&83.04&71.35\\
		\hline
		\multirow{3}{*}{\checkmark}&\checkmark&&\textbf{123.67}&80.97&\textbf{91.22}&\textbf{86.02}&204.24&99.12&\textbf{88.34}&\textbf{82.17} \\
		&&\checkmark&126.86&81.65&77.92&43.80&155.58&91.26&75.99&43.18\\
		&\checkmark&\checkmark&123.95&\textbf{80.39}&89.28&82.10&\textbf{154.89}&\textbf{91.19}&86.93&79.23\\
		\hline
	\end{tabular}
\end{table*}

\begin{table*}[!ht]
	\centering
	\caption{Ablation Study on Decoder.}
	\label{tab3}
	\begin{tabular}{c|cccc|cccc}
		\hline
		\multirow{2}{*}{Model}&\multicolumn{4}{c|}{Synthetic sketch}&\multicolumn{4}{c}{Freehand sketch} \\ \cline{2-9}
		&MPJPE$\downarrow$&Reconst. Error$\downarrow$&$\mathit{Acc.}$$\uparrow$&$\mathit{F1}$$\uparrow$&MPJPE$\downarrow$&Reconst. Error$\downarrow$&$\mathit{Acc.}$$\uparrow$&$\mathit{F1}$$\uparrow$ \\ \cline{1-9}
		\hline
		Single-branch&\textbf{114.54}&80.68&87.52&78.36&\textbf{147.35}&91.38&85.23&75.66 \\
		\hline
		Multi-branch&123.95&\textbf{80.39}&\textbf{89.28}&\textbf{82.10}&154.89&\textbf{91.19}&\textbf{86.93}&\textbf{79.23}\\
		\hline
	\end{tabular}
\end{table*}

\subsection{Comparison with state-of-the-art methods}

We evaluate our method by comparing with several state-of-the-art methods, including HMR \cite{kanazawa2018end}, SPIN \cite{kolotouros2019learning}, CMR \cite{kolotouros2019convolutional}, MAED \cite{wan2021encoder} and HybrIK \cite{oreshkin20233d}. We use SRT to train our model. For a fair comparison, the compared methods are trained following the same training procedure as our method. The SPIN and HMR use the SMPL parameter model regression, while the CMR directly regresses grid vertices and outputs its SMPL regression results. 

Table~\ref{tab1} shows the performance of our method and the comparisons. We also demonstrate the visual comparisons results in Fig.~\ref{fig4}. 

\subsubsection{Comparison on the synthetic sketch}
The ``synthetic skecthes'' columns in Table~\ref{tab1} show the evaluation results of image-model alignment and 3D regression on synthetic data. Our method achieves the best performance in all metrics. 

From the evaluation results, it can be observed that, our method improves 1.96\% accuracy and 4.1\% $\mathit{F1}$ score than MAED,
and achieves the best results in terms of 3D reconstruction performance, demonstrating the effectiveness of our method. It also implies that computing attention separately for $pose$, $shape$, and $cam$ is highly effective.
In contrast, the pure Transformer-based model HybrIK \cite{oreshkin20233d} fails to achieve good results.

\subsubsection{Comparison on the freehand sketch}
Our network exhibits good reconstruction performance from freehand sketches, as manifested in the ''freehand sketches'' columns in Table~\ref{tab1}. Our method performs best on freehand data, especially for $\mathit{F1}$ and $\mathit{Acc.}$, thanks to the multi-branch structure of our network and the advantages of hybrid training.
The visual comparison results can also be seen in Fig.~\ref{fig4}.
As shown in the last column of Fig.~\ref{fig4}, our method has great advantages in both posture and projection after being trained with the sketch3D dataset. 
Our method of using a deep residual network for sketch feature extraction and computing attention on the feature vectors with a separate attention component has shown significant improvement compared to previous methods that use a single-branch decoder or multiple channels without attention. 
This is because, when dealing with sketches, we need to take good account of both global information of key points and local features near adjacent key points. Using features extracted solely by CNNs overly emphasizes local information and loses some critical global information, while our attention mechanism can re-calculate the attention of each feature vector, globally enhance some critical information, and ultimately obtain a comprehensive understanding of both local and global information.

\subsection{Ablation Study}
We conduct ablation study on the training datasets, the networks with and without the attention module for both the synthetic (\textbf{Syn.}) and freehand (\textbf{Free.}) datasets, as well as the single-branch and multi-branch decoders. The training settings of these experiments are consistent with those in the implementation details.

\noindent\textbf{Ablation study on dataset.}~ 
To analyze the impact of the training data, we train our network using synthetic and freehand sketch datasets, respectively. The results are shown in rows 4, 5 and 6 in Table \ref{tab2}.
After SRT with freehand data, as shown in the fourth and sixth rows of Table~\ref{tab2} for Freehand sketch, the MPJPE and Reconst.Error of 3D reconstruction are reduced by 49.35 and 7.93, respectively. It demonstrates the importance of incorporating the freehand sketch dataset for training.

\noindent\textbf{Ablation study on attention.}~ 
In the ablation study of the attention module, the results are shown in the rows 3 and 6 of Table~\ref{tab2}. The result of w/o attention shows that the performance of our method is also significantly improved by such a mechanism. 
The module with attention is more accurate for the sketch-based 3D reconstruction of different shapes and can capture more information. 

\noindent\textbf{Ablation study on decoder.}~ 
We perform an ablation study on the decoder of the network. The results are reported in Table~\ref{tab3}.
Although the single-branch decoder has smaller errors than the multi-branch decoders in terms of MPJPE, this may be due to the fact that the 3D labels have large outliers. MPJPE is the mean squared error between the output and labels, and this calculation amplifies errors. On the other hand, Reconst.Error is the mean absolute error, which better reflects the 3D reconstruction error in the case of large outliers. Moreover, one important purpose of dividing into multiple branches is to better transfer the performance of predicting $cam$ from the \textbf{Syn.} dataset. As shown in Table~\ref{tab3}, the projection alignment performance has significantly improved on both \textbf{Syn.} and \textbf{Free.} datasets for multi-branch decoders.

\section{Conclusion}
In this work, we investigate a sketch-to-3D human reconstruction task and propose a multi-faceted decoder network SketchBodyNet.
Further, we build a large-scale Sketch3D dataset of freehand and synthetic sketches. It contains 26k freehand sketches of various styles and features the most diversity of human postures.
Extensive evaluation demonstrates the effectiveness of our method.

\section*{Acknowledgments}

This work was supported in part by GuangDong Basic and Applied Basic Research Foundation (No. 2021A1515012302, 2023A1515012845, 2023A1515011639, and 2022A1515011978), National Natural Science Foundation of China (No. 62272494), Key Special Project of Next-generation Electronic Information (Semiconductor) in Guangdong Province's ``Strong Innovative Universities Project'' (No. 2022ZDZX1007),
Guangdong Provincial Science and Technology Plan Project (No. STKJ2023069 and STKJ202209003), and Fundamental Research Funds for the Central Universities, Sun Yat-sen University (No. 23ptpy111).




\bibliographystyle{eg-alpha-doi}

\bibliography{egbibsample}

\end{document}